\documentclass[11pt,a4paper]{article}
\usepackage[hyperref]{acl2020}
\usepackage{times}
\usepackage{latexsym}

\usepackage{microtype}

\usepackage[ruled, lined, linesnumbered, longend]{algorithm2e} 

\usepackage{graphicx}
\graphicspath{{images/}}

\aclfinalcopy 
\def\aclpaperid{***} 

\title{Pretraining Federated Text Models for Next Word Prediction}

\author{%
  Joel Stremmel\thanks{\,Authors contributed equally to this work.} \\
  University of Washington\\
  Seattle, WA 98105 \\
  \texttt{jstremme@uw.edu} \\
  \And
  Arjun Singh\footnotemark[1] \\
  University of Washington \\
  Seattle, WA 98105 \\
  \texttt{arjuns13@uw.edu} \\
}

\begin{document}
\maketitle
\begin{abstract}
Federated learning is a decentralized approach for training models on distributed devices, by summarizing local changes and sending aggregate parameters from local models to the cloud rather than the data itself.  In this research we employ the idea of transfer learning to federated training for next word prediction (NWP) and conduct a number of experiments demonstrating enhancements to current baselines for which federated NWP models have been successful. Specifically, we compare federated training baselines from randomly initialized models to various combinations of pretraining approaches including pretrained word embeddings and whole model pretraining followed by federated fine-tuning for NWP on a dataset of Stack Overflow posts. We realize lift in performance using pretrained embeddings without exacerbating the number of required training rounds or memory footprint. We also observe notable differences using centrally pretrained networks, especially depending on the datasets used. Our research offers effective, yet inexpensive, improvements to federated NWP and paves the way for more rigorous experimentation of transfer learning techniques for federated learning.
\end{abstract}

\section{Introduction}

Machine learning on big data is an extremely popular and useful field of research and development. However, there are a variety of limitations to centrally aggregating data, such as compromised user privacy, single point of failure security risks, and the maintenance of often expensive hardware and compute resources. Federated learning aims to address this and has exhibited promising results for text completion tasks on mobile devices \citep{hard2018federated}. The Tensorflow Federated API provides methods to train federated models and conduct federated learning experiments on data grouped by clients but never aggregated \citep{bonawitz2019federated}. We build on the existing body of federated learning experiments, focusing on enhancing accuracy and reducing the required number of training rounds for federated text models for NWP through a variety of pretraining approaches.

\section{Related Work}

This research builds on related work in language modeling and federated learning to demonstrate the benefits of pretraining language models in the federated setting which are designed for next word prediction. We use an LSTM-RNN language model as in \citet{jing2019} for NWP, and take inspiration from the shallow LSTM \citep{LSTM2017} architecture with 10M parameters from \citet{melis2017}.  

\paragraph{}
For federated model training, we use the Federated Averaging Algorithm from \citet{McMahan2017} which averages model parameters after applying gradient updates to local models based on individual client datasets.  Our network is directly comparable to the network architecture used by \citet{reddi2020adaptive}\footnote{\url{https://github.com/tensorflow/federated/tree/master/tensorflow_federated/python/research/optimization/stackoverflow}} and similar to the federated RNNs in \citet{McMahan2017} and \citet{hard2018federated} in that we train on a dataset split by clients.  In this case, client datasets are collections of posts from Stack Overflow users, and we apply our language model to predict the next word in a given Stack Overflow post, similar to predicting the next word of a text message as in \citet{hard2018federated}.

\section{Enhancing Federated Text Models with Pretraining Methods}

 We apply three enhancements to federated training of our LSTM-RNN language model, demonstrating increased top-1 accuracy with fewer required training rounds. Our enhancements include:

\begin{enumerate}
  \item{Central pretraining followed by federated fine-tuning.}
  \item{Using a pretrained word embedding layer instead of randomly initialized embeddings during federated training.}
  \item{Combining centralized model pretraining and pretrained word embeddings with federated fine-tuning.}
\end{enumerate}

The following sections detail the methods we apply to achieve these enhancements as well as our experimental results. All code for this research is freely available under the MIT license in our GitHub repository\footnote{\url{https://github.com/federated-learning-experiments/fl-text-models}}.

\section{Data}

The main dataset used for these experiments is hosted by Kaggle and made available through the tff.simulation.datasets module in the Tensorflow Federated API \citep{bonawitz2019federated}. Stack Overflow owns the data and has released the data under the CC BY-SA 3.0 license. The Stack Overflow data contains the full body text of all Stack Overflow questions and answers along with metadata, and the API pointer is updated quarterly. The data is split into the following sets at the time of writing:

\begin{itemize}
\item{342,477 distinct users and 135,818,730 training examples}
\item{38,758 distinct users and 16,491,230 validation examples}
\item{204,088 distinct users and 16,586,035 test examples}
\end{itemize}

Challenges with the data include the size of the data and the distribution of words. As is common with text data (Zipf's law), the most common words occur with frequency far greater than the least common words. Therefore, in our experiments, we limit the vocab size to exclude very rare words. We provide a notebook of exploratory data analysis in our GitHub repository.

\paragraph{}
For the task of model pretraining, we also leverage the collected works of Shakespeare (as in the RNN from \citet{McMahan2017}) from Project Gutenberg released under the Project Gutenberg license \citep{shakespeare}. We download the full text of these collected works totaling 124,788 lines.

\section{Model Design}
In this study, we train a variety of small and large neural networks with four layers each as in table \ref{model-size-table}. 

\begin{table*}[ht]
\centering
\begin{tabular}{ccccc}
\hline \textbf{Size} & \textbf{Embedding Size} & \textbf{LSTM Size} & \textbf{Dense Layer Size} & \textbf{Output Layer Size} \\ \hline
Small & 100	& 256 & 100 & 10,004 \\
Large & 300	& 512 & 300 & 10,004 \\
\hline
\end{tabular}
\caption{\label{model-size-table} Model sizes.}
\end{table*}

\paragraph{}
The output layer represents the top 10,000 most frequently occurring vocab words in the Stack Overflow dataset plus four special tokens used during training denoting: padding, beginning of a sentence, end of a sentence, and out of vocabulary. We report accuracy with and without these tokens included.

\paragraph{}
We train both networks using the Adam optimizer and Sparse Categorical Cross Entropy loss for batches of size 16 and compare train and validation accuracy at each training round for 800 training rounds by sampling 10 non-IID client datasets per round, though we run some initial tests with 500 training rounds and a final test with 1,500. Each client dataset has 5,000 text samples from Stack Overflow at maximum, and a total of 20,000 validation samples. Model parameters are averaged centrally after each federated training round and the contribution of each client dataset to the Sparse Categorical Cross Entropy loss function is weighted by the number of text samples drawn from each client. We do not apply additional training rounds on the client datasets before averaging parameters and for this reason use the terminology of rounds and epochs interchangeably.

\paragraph{}
All models are trained with the Federated Averaging algorithm as in \citet{McMahan2017} using the Tensorflow Federated simulated training environment from \citet{bonawitz2019federated}. The large network outperforms the small network but with about three times the number of trainable parameters (7,831,328 vs 2,402,072) and is about three times the size (31.3MB vs 9.6MB). See the model layers in table \ref{model-size-table}.

\section{Central Pretraining with Federated Fine-Tuning}

The communication and computation costs of training models across distributed devices necessitates limiting the number of federated training rounds as much as possible. Transfer learning provides a way to trade computation time on independent devices for computation time on a central server. In this way, we propose that by initializing weights for a model to be trained on federated, private data with pretrained weights learned from centralized, public data, it is possible to limit training rounds on distributed devices, as the federated model will begin training with some information about the sequence of words, that is, which word should follow the text observed so far. We recognize that the English in Shakespeare differs greatly from the English in Stack Overflow posts, and therefore submit that the value of our work is mostly mechanical in nature, providing a simple method to extract weights learned from a centrally trained model and apply them to a model to be trained in the federated setting.

\paragraph{}
To centrally pretrain our federated model, we first load, preprocess, and fit a model to a pretraining dataset using the Keras submodule from Tensorflow. In doing so, we fit the same model architecture as described above for federated training but to the entire dataset for a predefined number of pretraining rounds. We then extract the tensors of model weights from the trained model and use these layer tensors to initialize the federated model. In the results to follow we pretrain either on Shakespeare or Stack Overflow.

\paragraph{}
For Stack Overflow, we use distinct samples for pretraining and fine-tuning to avoid overfitting and derive these samples from predefined splits of the Stack Overflow data from the data loading API mentioned previously.  We pretrain on the set labeled "test" and report validation performance on the set labeled "validation," fine-tuning on the set labeled "train." Words that do not map to embeddings learned during pretraining are initialized by drawing floating points from the random uniform distribution on the interval [-0.05, 0.05]. We apply this same method of filling in missing words when using pretrained word embeddings for federated training on Stack Overflow which we describe in the next section.

 \begin{figure*}[ht]
 \centering
  \includegraphics[width=0.8\textwidth]{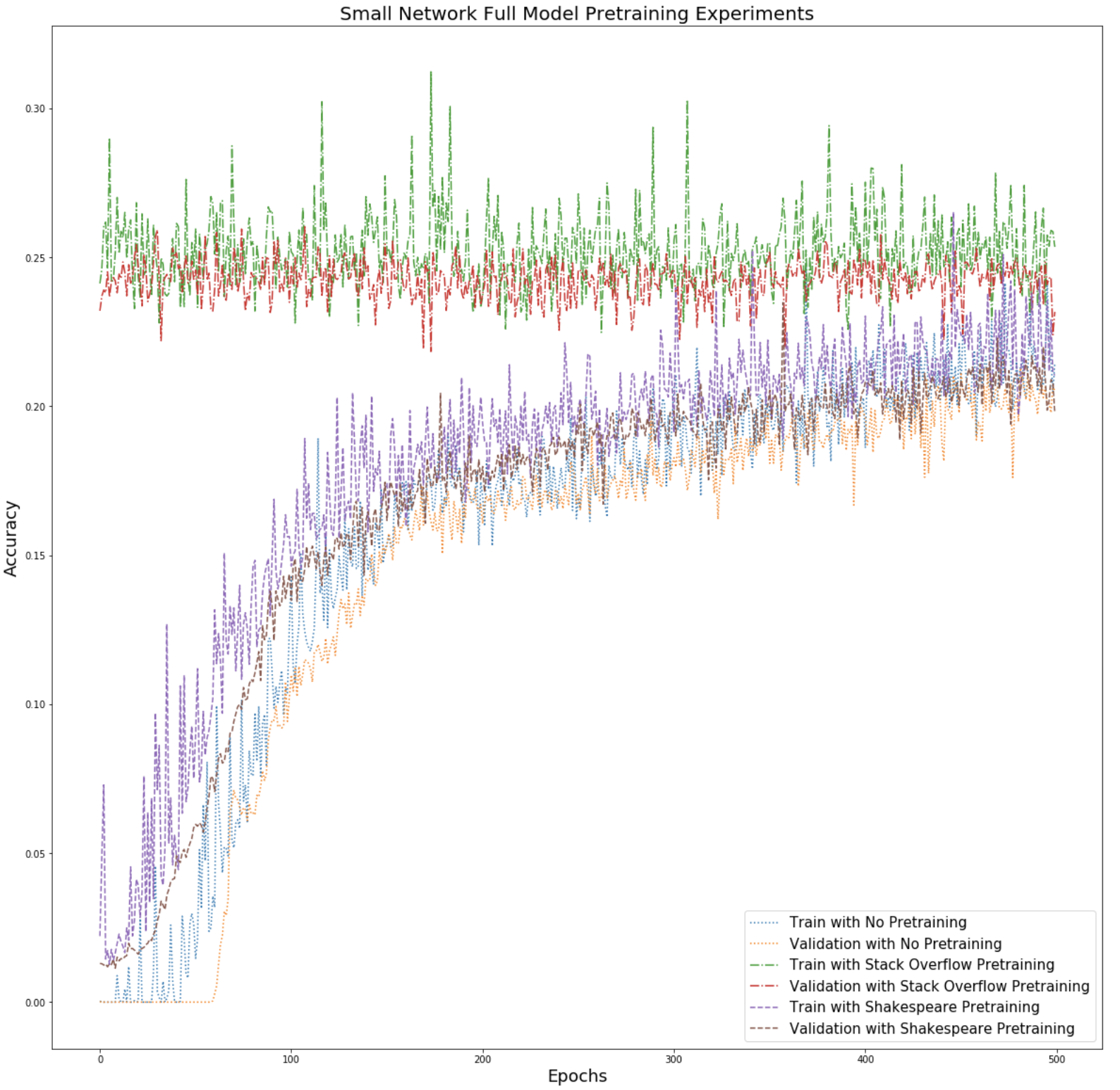}
 \caption{\label{pretrain-fig} Model pretraining experiments with Stack Overflow and Shakespeare.}
 \end{figure*}

\paragraph{}
We fine-tune three different models in the federated style for 500 rounds (figure \ref{pretrain-fig}). Although the network remains the same for all three, the key difference is whether they are pretrained. The three models are as follows:

\begin{enumerate}
  \item{Federated training on Stack Overflow without any pretraining which yields the two learning curves that exhibit the lowest levels of train and validation accuracy respectively.}
  \item{Central pretraining on Shakespeare for 50 rounds followed by federated fine-tuning on Stack Overflow which yields curves exhibiting marginal lift in both train and validation accuracy.}
  \item{Pretraining on distinct Stack Overflow IDs with federated fine-tuning.}
\end{enumerate}

\paragraph{}
The two main takeaways from this experiment are as follows:
\begin{enumerate}
	\item{Pretraining generally improves the performance of fine-tuning.}
	\item{When the source of data is identical for pretraining and fine-tuning, fine-tuning adds no value.}
\end{enumerate}

\paragraph{}
We must also note that for the latter case, it may not be practically possible to have the same source of data for pretraining, performed centrally on a server, and fine-tuning, performed in the federated style on user devices.

\section{Pretrained Word Embeddings for Federated Training}

We hypothesize that having a common, starting representation for words across federated (non-IID) datasets yields improved model performance with fewer training rounds compared to federated training with randomly initialized word embeddings. To test this, we consider a variety of pretrained word embeddings including GloVe \citep{pennington2014glove}, FastText \citep{bojanowski2016}, and GPT2 \citep{radford2019language} for both our small and large network architectures. These methods of pretraining word embeddings vary in implementation, capturing different information about word relationships. In practice each embedding method exposes a preselected vocabulary with vector representations for each word, and can thus be compared on the basis of how these vector representations enable various downstream tasks. For the present task of NWP, we expect the GPT2 embeddings, trained in an autoregressive fashion for NWP, to encode especially relevant information for our task of predicting the next word in Stack Overflow posts. We retrieve GPT2 embeddings from the HuggingFace Transformers Library \citep{Wolf2019HuggingFacesTS}.

\paragraph{}
While GloVe embeddings are commonly used and come in a variety of dimensions (50, 100, 200, 300), FastText and GPT2 embeddings are limited to a handful of sizes. We test the 100 and 300-dimensional GloVe embeddings in the small and large networks respectively and the 300-dimensional FastText embeddings in the large network. To create 100-dimensional FastText embeddings as well as 100 and 300-dimensional GPT2 embeddings from the smallest available GPT2 embeddings of size 768, we use two methods:

\begin{enumerate}
 \item{We apply Principal Components Analysis to reduce these word embeddings to the desired dimensions of 100 and 300 and include these word embeddings in our experiment runs.}
 \item{We run the same experiments but achieve 100 and 300-dimensional FastText and GPT2 embeddings using algorithm \ref{algo2}, Principal Components Analysis with the Dimensionality Reduction Algorithm from \citet{raunak-etal-2019-effective}.}
\end{enumerate}

\paragraph{}
The Dimensionality Reduction Algorithm applies algorithm \ref{algo1}, the post-processing algorithm from \citet{mu2018allbutthetop}, which subtracts the mean vector from all word vectors as well as the directions of variation explained by the top D principal components.

\begin{algorithm}
    \KwData{Word Embedding Matrix X, Threshold Parameter D}
    \KwResult{Post-Processed Word Embedding Matrix X}

    \tcc{Subtract Mean Embedding}
    $X = X - \bar{X}$

    \tcc{Compute PCA Components}
    $u_{i} = PCA(X)$ where $i=1, 2, \ldots, D$

    \tcc{Remove Top-D Components}
    \For{all v in X}{
    	$v = v - \sum_{i=1}^{D} (u_{i}^{T} \cdot v) u_i$
    }
    \caption{\label{algo1} Post-Processing Algorithm PPA(X, D)}
\end{algorithm}

\paragraph{}
The intuition behind algorithm \ref{algo1} is that the mean vector for a set of word embeddings as well as the dominating principal components describe variation that is common across the embedding space, and therefore do little to help distinguish between individual word representations. While \citet{mu2018allbutthetop} demonstrate that the post-processing algorithm yields improved performance on a variety of word similarity tasks by purifying word embedding representations, \citet{raunak-etal-2019-effective} demonstrate the benefits of applying post-processing before and after dimensionality reduction via principal components (algorithm \ref{algo2}) through improved performance on some word similarity benchmarks compared to the post-processing algorithm alone (algorithm \ref{algo1}), while across word similarity benchmarks achieving at least equal performance on a majority of tasks but with significantly smaller embeddings.

\begin{algorithm}
    \KwData{Word Embedding Matrix X, New Dimension N, Threshold Parameter D}
    \KwResult{Word Embedding Matrix of Reduced Dimension N: X}

    \tcc{Apply Algorithm 1 (PPA)}
    $X = PPA(X, D)$

    \tcc{Transform X Using PCA to N Dimensions}
    $X = PCA\_Transform(X)$

    \tcc{Apply Algorithm 1 (PPA)}
    $X = PPA(X, D)$

    \caption{\label{algo2} Dimensionality Reduction Algorithm PP\_PCA\_PP(X, N, D)}
\end{algorithm}

\paragraph{}
This dimensionalty reduction approach is useful for federated training in which we are constrained by model size, and we use this approach to create word embeddings when embeddings of our desired sizes (100 and 300) are not available.  We use these embeddings for federated NWP with the aforementioned model architectures.  In the plots to follow "PCA" indicates the use of word embeddings reduced by PCA transformation, while "PP PCA PP" indicates the use of algorithm \ref{algo2} with D=7 (as in \citet{raunak-etal-2019-effective} but also based on plotting variance explained for our word vectors).  We measure train and validation accuracy with end of sentence and out of vocab tokens over 800 rounds and report test accuracy with and without these tokens in table \ref{test-set-table} by freezing model weights at the training round achieving the best validation accuracy. We omit the small network validation accuracy plots across our variety of word embedding representations (available in our GitHub repository\footnote{\url{https://github.com/federated-learning-experiments/fl-text-models/blob/master/final_research_report/images/small_emb_results_grid.png}}) but show test set performance in table \ref{test-set-table} for both the small and large network experiments.

\begin{figure*}[!ht]
 \centering{}
  \includegraphics[width=0.8\textwidth]{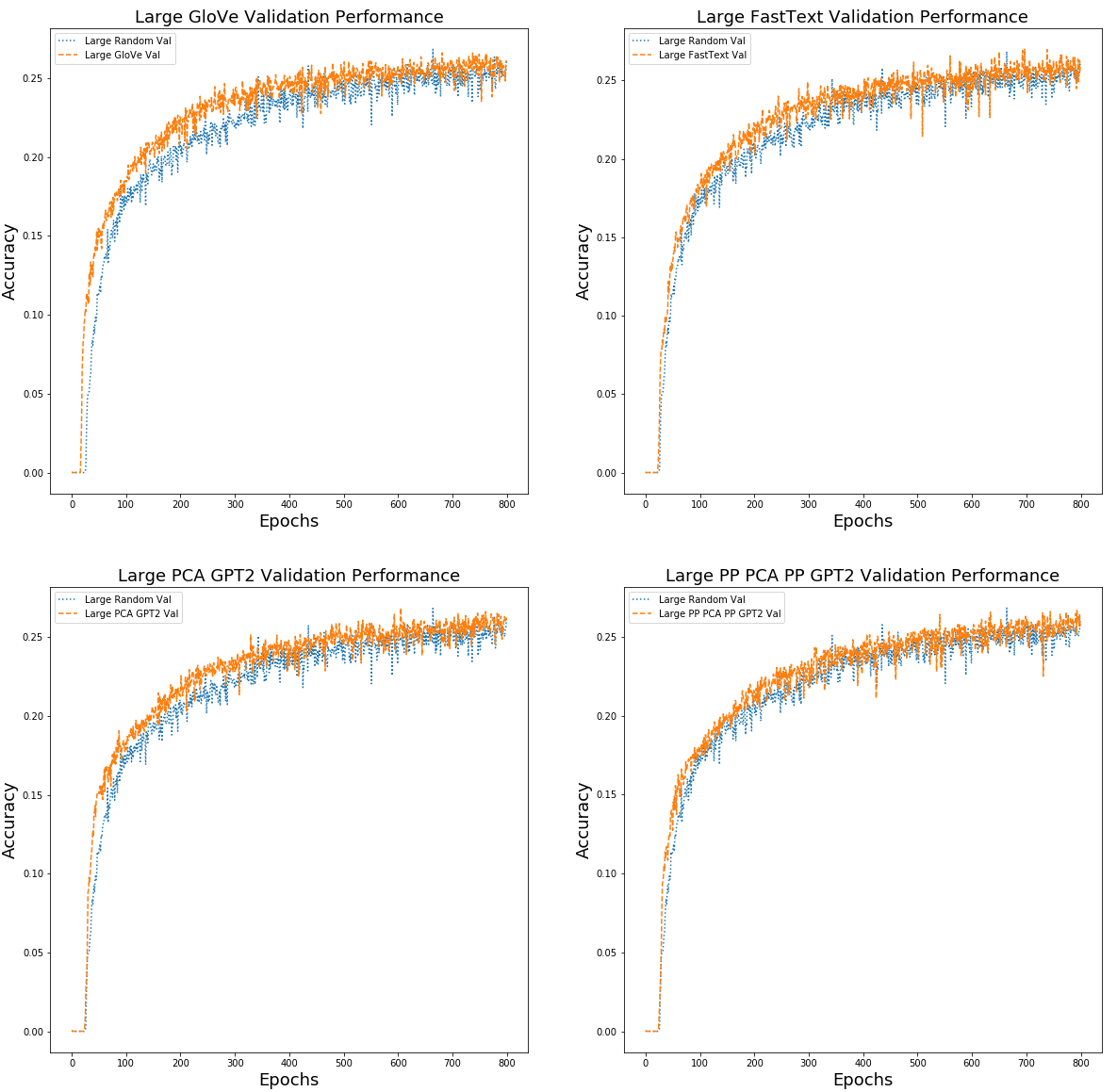}
 \caption{\label{large-emb-perf-fig} Grid of pretrained word embedding layer results compared to random embeddings for large networks.}
\end{figure*}

\paragraph{}
In the small networks, the GloVe embeddings start to gain accuracy ahead of all other approaches, while the randomly initialized embeddings require more training rounds to achieve the same level of accuracy early on in the training process compared to pretrained word embeddings. These benefits of pretraining are more pronounced in the large networks as in figure \ref{large-emb-perf-fig}, where pretrained word embeddings achieve the same level of accuracy sooner, that is, with fewer training rounds compared to random embeddings. This early boost in performance is valuable in the federated setting in the sense that these embeddings will take up no more space than random embeddings and help the model approach peak accuracy with fewer training rounds, each of which requires communication between the server averaging model parameters and the training clients.

\paragraph{}
Comparing the models trained with these word embeddings on 1,000,000 text samples from the Stack Overflow test set in table \ref{test-set-table}, we observe an increase of over half a percent accuracy with pretrained compared to random embeddings for the large networks with little to no improvement from pretrained embeddings for the small networks. We highlight the large network GPT2 word embeddings with reduced dimension via the Dimensionality Reduction Algorithm as the best performing approach in terms of accuracy, both with and without end of sentence and out of vocab tokens.  

\begin{table*}[ht]
\centering
\begin{tabular}{lcccc}
\hline \textbf{Model} & \textbf{Accuracy} & \textbf{Accuracy No OOV No EOS} & \textbf{Parameters} & \textbf{Weights(MB)}\\ \hline
\textbf{Small Random*} & \textbf{0.2246} & \textbf{0.1821} & 2.4M & 9.6 \\
Small GloVe & 0.2269 & 0.1838 & 2.4M & 9.6 \\
Small PCA FastText & 0.2250 & 0.1823 & 2.4M & 9.6 \\
Small PP\_PCA\_PP FastText & 0.2285 & 0.1852 & 2.4M & 9.6 \\
Small PCA GPT2 & 0.2293 & 0.1859 & 2.4M & 9.6 \\
Small PP\_PCA\_PP GPT2 & 0.2262 & 0.1834 & 2.4M & 9.6 \\
\textbf{Large Random*} & \textbf{0.2485} & \textbf{0.2086} & 7.8M & 31.3 \\
Large GloVe & 0.2557 & 0.2162 & 7.8M & 31.3 \\
Large FastText & 0.2548 & 0.2137 & 7.8M & 31.3 \\
Large PCA GPT2 & 0.2522 & 0.2118 & 7.8M & 31.3 \\
\textbf{Large PP\_PCA\_PP GPT2**} & \textbf{0.2569} & \textbf{0.2169} & 7.8M & 31.3 \\
\hline
\end{tabular}
\caption{\label{test-set-table} Model performance by embedding layer experiment. *Baseline approach. **Best performing.}
\end{table*}

\section{Federated Fine-Tuning Using a Pretrained Model with Pretrained Word Embeddings}

\begin{figure}[ht]
 \centering{}
  \includegraphics[width=0.4\textwidth]{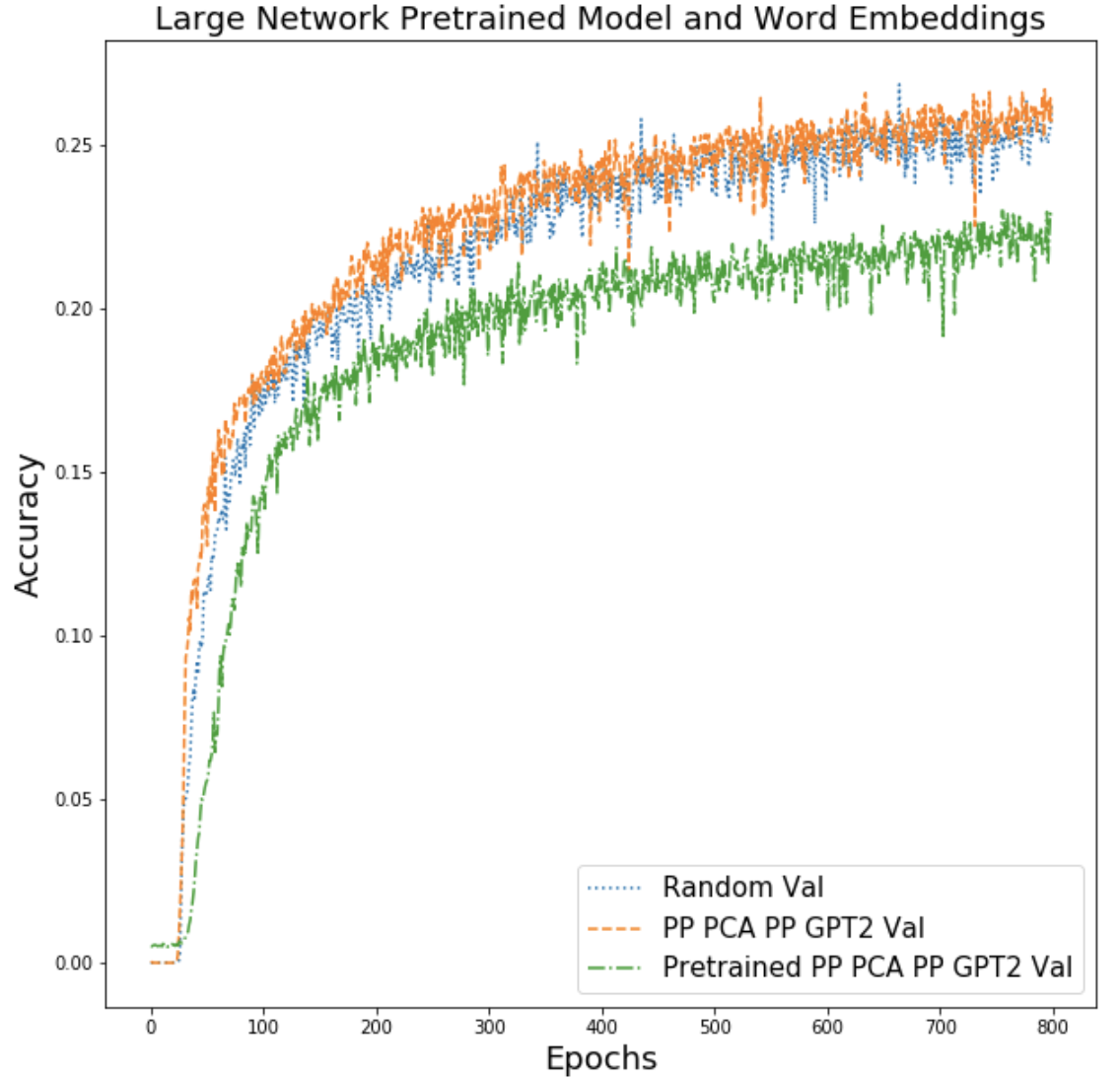}
 \caption{\label{combined-fig} Large network pretrained model and word embeddings compared to no pretraining.}
\end{figure}

As both model pretraining and starting with pretrained word embeddings provide ways of kicking off federated training with more intelligent models, it is natural to combine the two approaches. In doing so we observe that even with the best of our word embedding approaches, the pretrained model (50 pretraining rounds with 800 rounds of fine-tuning) performed worse than starting with federated training using both random and pretrained embeddings (figure \ref{combined-fig}).

\paragraph{}
We suspect that while pretraining with Shakespeare is effective for the small network, using a model with increased capacity renders this prior information useless, as the nature of Shakespearean English is quite different from that of Stack Overflow. In this way we think that a dataset more similar to Stack Overlow may yield increased performance for full model pretraining.

\section{Comparison to Adapative Federated Averaging Stack Overflow Baseline}

\begin{figure}[ht]
 \centering{}
  \includegraphics[width=0.4\textwidth]{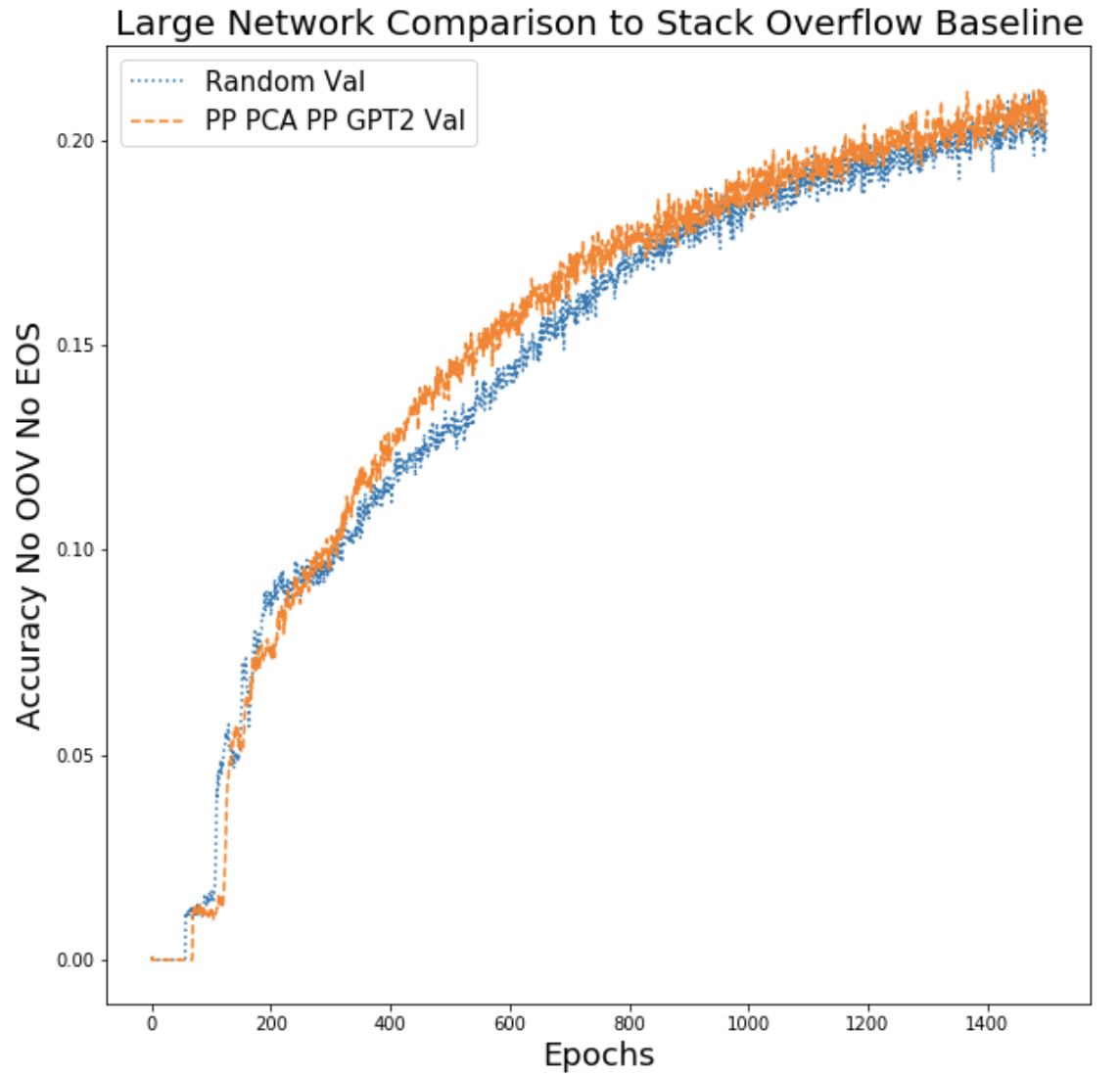}
 \caption{\label{adptv-curves-fig} Random and reduced GPT2 embedding comparison. Training configuration from \citet{reddi2020adaptive}.}
\end{figure}

Our pretraining experiments fixed the client sample size and model architecure as described earlier, though to demonstrate robustness, we explore whether or not the successes we observe with pretraining, particularly using pretrained word embeddings with the Dimensionality Reduction Algorithm, will still hold with a different federated client sample size and model architecure. In \citet{reddi2020adaptive}, the authors sample 50 clients per training round with a max of only 128 text samples instead of 5,000 as in our experiments. They also use an embedding dimension of size 96 with an LSTM layer of size 670, feeding to two dense layers of size 96 and 10,004 respectively. With this approach we compare randomly initialized word embeddings to our best performing pretrained word embeddings: reduced GPT2 embeddings via algorithm \ref{algo2}.  See the learning curves in figure \ref{adptv-curves-fig} and final evaluation in table \ref{avg-table}.

\begin{table*}[t]
\centering
\begin{tabular}{lc}
\hline 
\textbf{Model} & \textbf{Accuracy No OOV No EOS} \\ 
\hline
Random & 0.2019 \\
Large PP\_PCA\_PP GPT2 & 0.2065 \\
\hline
\end{tabular}
\caption{\label{avg-table} Average performance over last 100 validation rounds. Training configuration from \citet{reddi2020adaptive}.}
\end{table*}

\paragraph{}
We find that pretrained word embeddings generally outperform random embeddings across 1,500 rounds of training with evaluation on 10,000 validation samples per training round and a final evaluation performed by averaging the last 100 rounds of validation accuracy without special tokens. While we demonstrate improvement over this baseline architecture using the same training and evaluation design from \citet{reddi2020adaptive}, we do not realize the same level of accuracy as the paper which achieves 22.1\%, and 22.2\% with Adam and Yogi optimizers respectively, as in our experiments we use only the default learning rates for Adam. Future work would apply adapative learning rate methods as in \citet{reddi2020adaptive} to both embedding approaches to see if pretrained embeddings continue to outperform random.

\section{Future Work}
While our initial research demonstrates the possibility of reducing the number of federated training rounds required to achieve acceptable model accuracy through the use of pretrained word embeddings, there is much left to explore. For central pretraining with federated fine-tuning, we demonstrate a viable procedure but do not achieve performance greater than the federated training baseline with our large network. That said, this approach may be fruitful with pretraining data more similar to Stack Overflow than the collected works of Shakespeare. Also, for both model pretraining and pretrained word embedding approaches, the adaptive learning rate method as in \citet{reddi2020adaptive} may help address the specific optimization requirements of fine-tuning weights that have already undergone some training. Additionally, using federated simulation to conduct pretraining, such that the initial model weights are learned on non-IID datasets, may improve overall model performance after federated fine-tuning. Simulating federated training conditions to pretrain word embeddings may also yield improved downstream performance by tailoring word representations to reflect different usage across non-IID datasets. 

\section{Conclusion}
While GPT2 and other Transformer-based models are achieving state of the art performance on centralized language modeling tasks, the sizes of these models are prohibitively large for federated training and prediction. Meanwhile, leveraging the embeddings learned from these models for federated training can be immediately useful. Our current research offers effective yet inexpensive improvements to federated models for NWP, particularly through the use of pretrained word embeddings, and also paves the way for more rigorous transfer learning experiments for federated learning.

\bibliography{acl2020}
\bibliographystyle{acl_natbib}

\end{document}